\documentclass[conference]{IEEEtran}
\IEEEoverridecommandlockouts
\usepackage{array}
\usepackage{booktabs} 
\usepackage{hyperref}
\hypersetup{
    colorlinks=false,   
    pdfborder={0 0 0}   
}
\usepackage{enumitem}
\usepackage{hyperref}
\usepackage{xcolor}
\usepackage{soul}
\sethlcolor{yellow}   

\usepackage{csquotes}
\usepackage{cite}
\usepackage{amsmath,amssymb,amsfonts}
\usepackage{array}
\usepackage{textcomp}
\usepackage{xcolor}
\usepackage{amsmath}
\usepackage{tikz}
\usepackage{nth}
\usepackage{subcaption} 
\usepackage{float}

\usepackage{algorithmicx,algpseudocode}
\usepackage{polyglossia}
\usepackage[linesnumbered, ruled, vlined]{algorithm2e}

\usepackage{siunitx,array,multirow}

\usepackage{tabularx}
\usepackage{multirow}
\usepackage{threeparttable}
\usepackage{multicol}
\usepackage{xcolor,colortbl}
\usepackage{amsmath}

\usetikzlibrary{shapes,arrows}
\usepackage{verbatim}

\usepackage{url}
\usepackage{amsmath}
\usepackage{textcomp}
\usepackage{siunitx}
\usepackage[utf8]{inputenc}
\usepackage{upgreek}

\makeatletter
 \let\old@ps@headings\ps@headings
 \let\old@ps@IEEEtitlepagestyle\ps@IEEEtitlepagestyle
 \def\confheader#1{%

 \def\ps@IEEEtitlepagestyle{%
 \old@ps@IEEEtitlepagestyle%
 \def\@oddhead{\strut#1\hfill\strut}%
 \def\@evenhead{\strut\hfill#1\hfill\strut}%
 }%
 \ps@headings%
 }
 \makeatother

 \confheader{
    \noindent  
    \fontsize{10}{11}\selectfont  
    \parbox{\textwidth}{ 
        2025 28th International Conference on Computer and Information Technology (ICCIT) \\ 
        19-21 December 2025, Cox’s Bazar, Bangladesh
    }
}

\usepackage[pscoord]{eso-pic}
\newcommand{\placetextbox}[3]{
\setbox0=\hbox{#3}
\AddToShipoutPictureFG*{ \put(\LenToUnit{#1\paperwidth},\LenToUnit{#2\paperheight}){\vtop{{\null}\makebox[0pt][c]{#3}}}
}
}
\placetextbox{.338}{0.08}{\small{Authors' version, published version's DOI: 10.1109/ICCIT68739.2025.11490128}}
    
\begin{document}
\title{Towards Developing a Multimodal Chat Assistant for University Stakeholders: RAG-based Approach}

\author{\IEEEauthorblockN{Md Abu Hanif Shaikh, and Abdullah Al Shafi}
\IEEEauthorblockA{
Institute of Information and Communication Technology, \\Khulna University of Engineering \& Technology, Khulna, Bangladesh\\
hanif@kuet.ac.bd, and abdullah@iict.kuet.ac.bd}}


\maketitle
\begin{abstract}
University stakeholders often face difficulties in accessing timely and reliable information, especially in developing countries, where there are very few intelligent support systems. Existing rule-based chatbots are unable to handle complex, domain-specific queries and are not well-equipped to adapt to evolving institutional policies. As a fill-in-the-gap solution, we present the multimodal university chatbot with retrieval-augmented generation. The system combines the large language model with semantic retrieval to produce context-based responses from institution-centric resources, such as the university handbook. The system accepts text and image queries through the vision-language model and applies quantized inference for rapid deployment on constrained hardware. A scalable backend built with FastAPI, adjoined with a responsive frontend developed with Next.js, ensures real-time usability. Our multimodal evaluation demonstrates that the system maintains strong satisfaction scores across both text and image queries, despite increased response time for visual inputs. Furthermore, quantitative evaluation shows that hallucination is reduced from 31.7\% to 6.6\% in our proposed RAG-based system, confirming the effectiveness of retrieval grounding.

\end{abstract}

\begin{IEEEkeywords}
Retrieval Augmented Generation, Vision Language Model, University Chat Assistant, Large Language Model, Vector Store
\end{IEEEkeywords}

\section{Introduction}
\label{sec:introduction}
Instead of searching through lengthy documents, people generally prefer to receive an instant and summarized response \cite{mcgrath2025generative}. That is why chatbots are getting popular day by day in various sectors like education, clinical decision support system, customer service, etc \cite{peyton2025review}. The same holds for universities and their stakeholders like teachers, administrative officers, students, staff, and potential applicants. University chatbots are very common in developed countries like the USA, UK, Canada, but almost unheard of in developing countries like Bangladesh \cite{peyton2025review}. Hence, building such a chatbot is indeed an important step toward the modernization of academic support infrastructure, bridging the digital divide, and enhancing student access to information in resource-constrained educational environments.


Furthermore, traditional rule-based chatbot systems and static knowledge bases frequently cannot handle complex queries involving subtle nuances in institutional policies, dynamic changes in course offerings, or personalized responses tied to user context \cite{peyton2025review}. This gap necessitates the urgent development of intelligent systems capable of reasoning, retrieval, and autonomous action.


Our proposed system addresses this challenge by leveraging retrieval-augmented generation (RAG) \cite{li2025retrieval}, an advanced paradigm in AI that integrates large language models (LLMs) with retrieval mechanisms. This approach enables the chatbot to retrieve relevant university-specific documents (e.g., academic policies, course catalogs, FAQs), which may change dynamically, without a computationally expensive fine-tuning process. The core of our system uses RAG to find and combine information from documents, giving accurate answers based on context. The backend uses FastAPI for handling many requests at once, and the frontend ensures a smooth user experience with Next.js and React. The main contributions of the work are listed below:

\begin{enumerate}[label=\roman*.]
    \item Development of a RAG-powered university chat assistant that processes both text and image queries utilizing a vision language model (VLM).
    \item To support this chat assistant, a retrieval pipeline utilizing semantic embeddings and chromaDB was constructed to provide context-aware, accurate responses from institutional documents.
    \item Implementation of quantized inference and asynchronous FastAPI backend for real-time, resource-efficient operation.
    \item Deployment of our proposed system as a publicly accessible chat assistant at \href{https://chat.kuet.ac.bd}{\texttt{chat.kuet.ac.bd}} with an interactive interface, feedback system, and demonstrated ability to handle hallucination issues.
\end{enumerate}

The rest of the paper is organized as follows, Section \ref{sec:literature} compares some existing works with limitations in the context of Bangladesh. The architecture of our suggested RAG-based system is described in Section \ref{sec:methodology}. The experimental setup and simulation are presented in Section \ref{sec:result}. The ethical issues are discussed in Section \ref{sec:privacy}. Finally, a conclusion is drawn in Section \ref{sec:conclusion}.

\begin{figure*}[htbp]
\centering
\centerline{\includegraphics[scale=0.46]{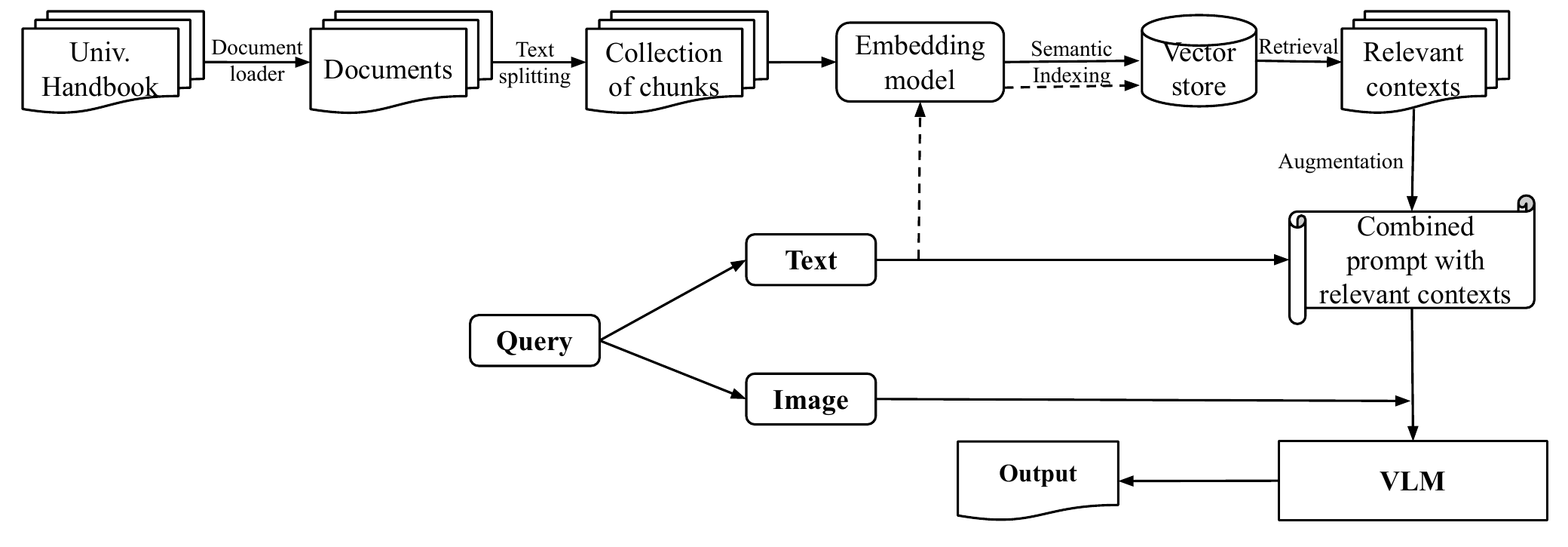}}
\caption{Architecture of our developed system.}
\label{rag}
\end{figure*}

\section{Related work}
\label{sec:literature}
Research on university chatbots has developed from early research on rule-based or semantically oriented models into advanced AI and hybrid architectures that increase levels of student support and streamline administrative tasks and academic resource accessibility.

At the very beginning, the paper \cite{ranoliya2017chatbot} proposes a university FAQ chatbot using AIML and latent semantic analysis to provide efficient, interactive, and accurate responses to student queries. After that, Patel et al. present UNIBOT \cite{patel2019ai}, a web-based university chatbot developed with a custom algorithm to provide fast, human-like responses to student queries. Using the Rasa framework and BERT and RNN models with an accuracy of 97.1\% in identifying intention, Nguyen et al. \cite{nguyen2021neu} developed an AI assistant for admission processes and therefore reduced the workload of admission personnel.


In the study of Balderas et al.\cite{balderas2023chatbot}, a chatbot system was developed to facilitate communication with university students in emergency situations, thus ensuring greater campus safety. The study \cite{attigeri2024advanced} reports the development and comparative study of five NLP-based chatbot models for counseling on university campuses, showing that sequential neural network models outclass TF-IDF and pattern-matching approaches.

Parrales-Bravo et al. \cite{parrales2024csm} have developed machine learning and NLP techniques to create a chatbot that is deployed in Telegram, assisting students of the University of Guayaquil in queries regarding enrollment and payment. It shows extensive usability and positive acceptance among the student population.

In \cite{li2025retrieval}, a hybrid chatbot integrating rule-based, retrieval-based, and generative approaches is proposed for administrative support in education, achieving higher accuracy and user satisfaction than stand-alone models. Neupane et al. \cite{neupane2024questions} present BARKPLUG V.2, a RAG-based chatbot that improves access to university resources by delivering accurate, context-specific answers, achieving high performance and satisfactory usability. The paper \cite{han2025mobile} introduces EduChatGPT, a personalized AI chatbot for education that integrates GPT-4 with real-time student data to deliver tailored learning support. It employs a modular architecture combining natural language processing, data integration, and adaptive feedback mechanisms. Experimental evaluation demonstrates improvements in student engagement and personalized learning outcomes. The work highlights the potential of generative AI to enhance individualized academic support in higher education contexts.

To the end, though there are several university chatbots leveraging various techniques for developed countries, there is a scarcity of such system in the developing and under-developed countries including Bangladesh. We have tried to mitigate the gap by designing a web-based chat application for university stakeholders in the context of Bangladesh.

\section{Proposed RAG-based System}
\label{sec:methodology}
As shown in Fig.\ref{rag}, an AI system was designed to accept multimodal input (text and image) to generate coherent and context-sensitive responses. The system utilizes cutting-edge vision language models, embedding techniques, and document retrieval techniques. Its foremost purpose is the understanding of queries posed in natural language or image form, and then providing a descriptive response by either querying from pre-defined knowledge sources (University handbook) or image analysis on a real-time basis.

\subsection{System Overview}
The system is configured as a web application with RESTful APIs support. It receives input from the user in the form of natural-language text or an image with text and dynamically selects the appropriate pipeline depending on the content of the input. The architecture has three main modules: (1) inference engine; (2) knowledge retrieval and augmentation; and (3) feedback collection and monitoring.


\begin{figure*}
     \centering
     \begin{subfigure}[b]{0.46\textwidth}
         \centering
         \includegraphics[width=\textwidth]{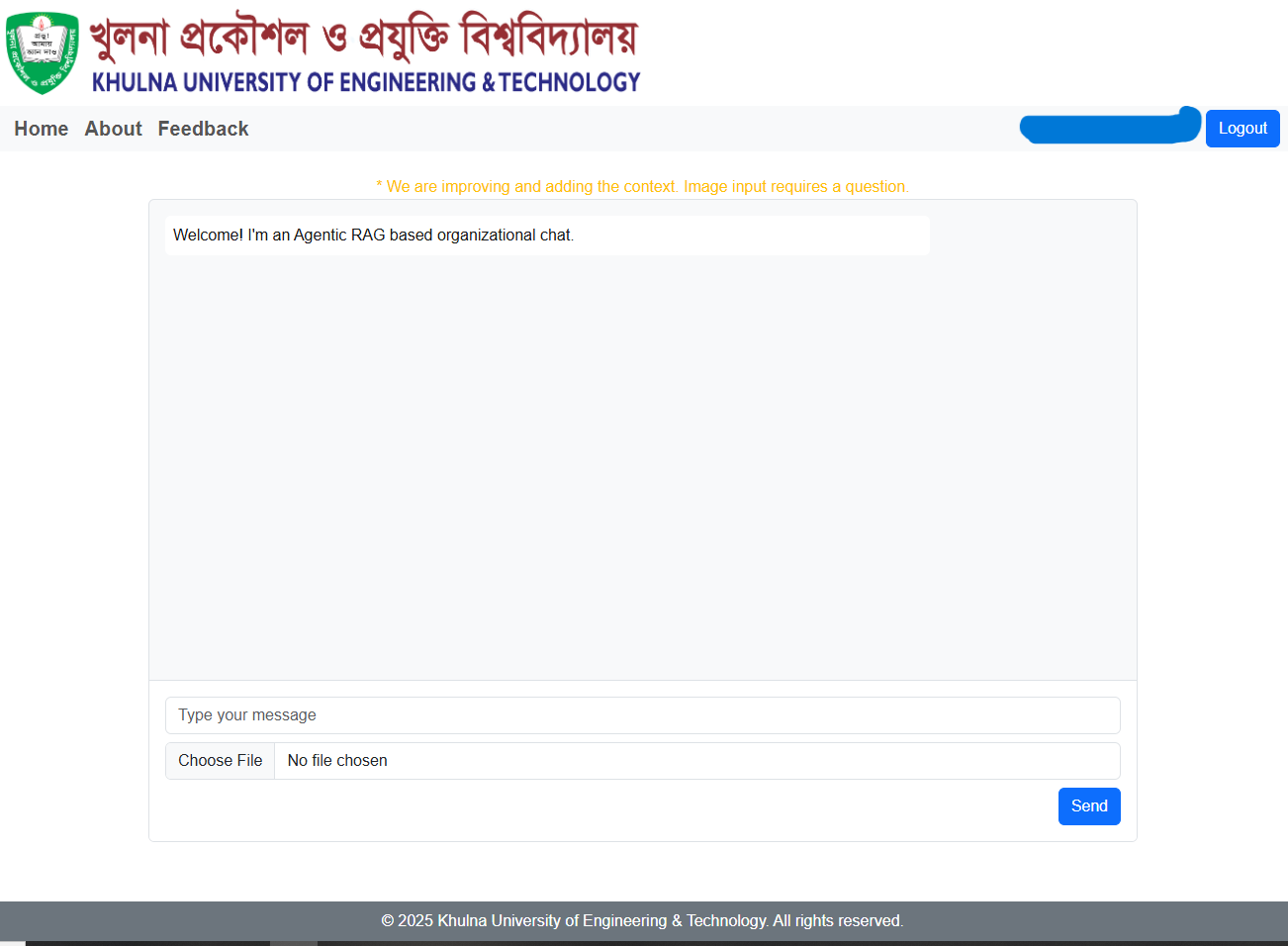}
         \caption{Chat interface}
         \label{base_ui}
     \end{subfigure}
     \hspace{1em}
     \begin{subfigure}[b]{0.46\textwidth}
         \centering
         \includegraphics[width=\textwidth]{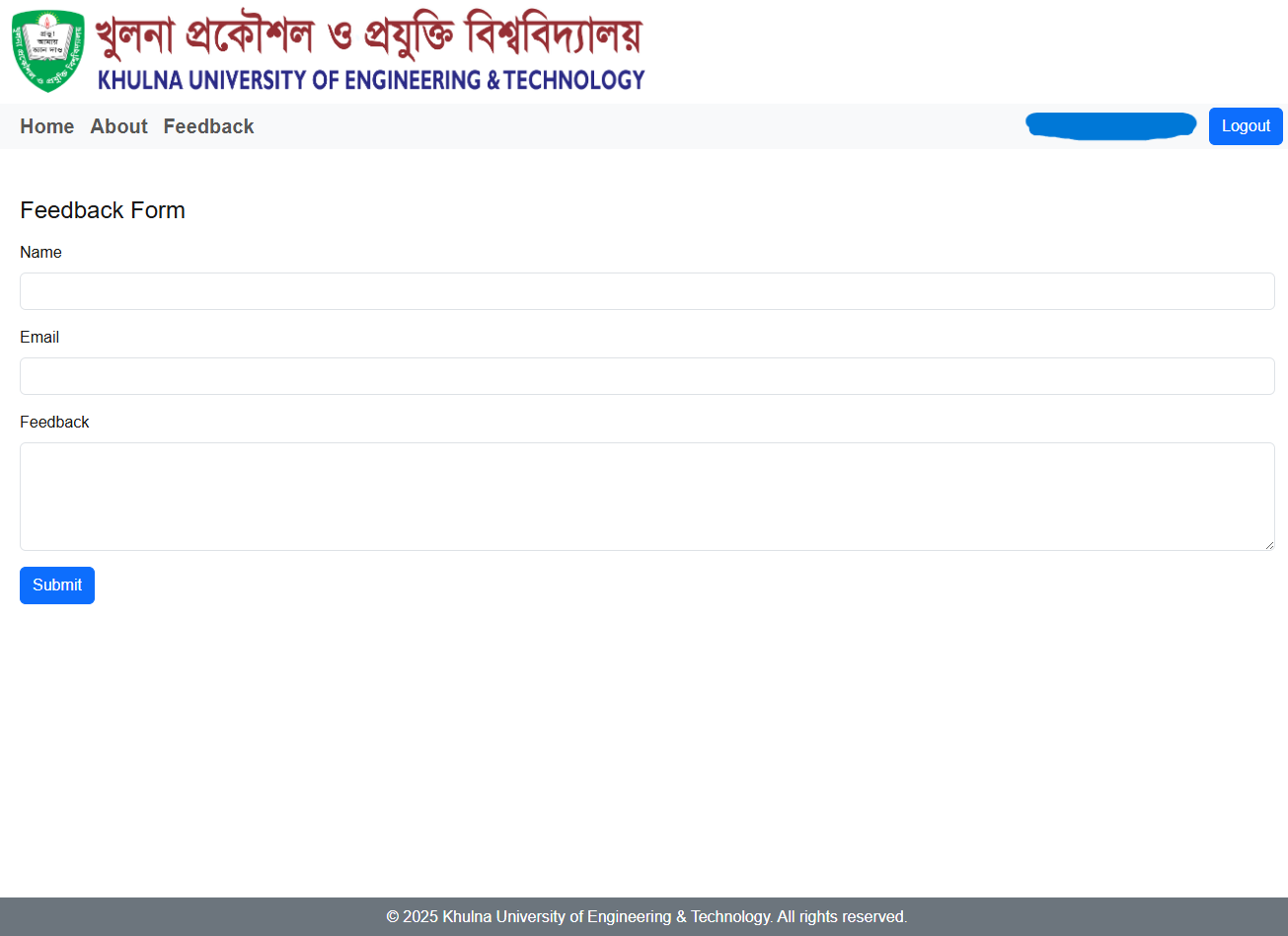}
         \caption{Feedback form}
         \label{feedback_form}
     \end{subfigure}
        \caption{Interactive user interface of our system.}
\label{interactive_ui}
\end{figure*}

\subsubsection{Multimodal Inference Engine}
The user inputs are processed by a large-scale VLM llava-1.5-7b \footnote{https://huggingface.co/liuhaotian/llava-v1.5-7b}, which is capable of multimodal processing. The model is loaded on a quantized level to minimize memory consumption and to make the inference faster, while maintaining sufficient accuracy for the description and interpretation tasks. A user may send either an image or a textual query, or both at the same time. The system, in case both modalities exist, builds a structured prompt incorporating the user's query and the image for context-aware visual reasoning. Text-only queries undergo language-only processing, with an optional step of supplementing in-context information from a domain-specific knowledge base. Whatever the scenario might be, the model excels at instruction following and natural language generation.

\subsubsection{Knowledge Base Construction}
To enhance the model’s responses with domain-specific knowledge, a static document corpus is preprocessed and indexed. The pipeline includes:

\textbf{Document loading: }The first step of the system concerns the ingestion of documents, where raw textual data is extracted from the knowledge source. In the consideration of this study, the official KUET university handbook, in PDF format, was taken as the primary source. Each page of the document was parsed to obtain line-level text. To reduce noise, regular expressions were applied to remove irrelevant elements such as roman-numeral headers, numeric indices, page markers, etc., none of which have semantic information at all. In this way, only meaningful content was retained, thus maximizing the relevance and accuracy of the other stages of the RAG pipeline.


\textbf{Text splitting: }Since LLMs are unable to process arbitrarily long sequences of text, the loaded document was thus divided into smaller, semantically coherent units. At first, the text is recursively attempted to be split at logical boundaries (paragraphs, sentences) before considering character-based splits. A chunk size of 512 characters, with an overlap of 64, is ideal for contextual continuity between adjacent chunks; ensuring that important phrases near the boundaries are shared in multiple chunks and therefore minimize information loss during retrieval. This balance between granularity and semantic integrity guarantees that chunks are rich enough contextually and small enough computationally to be handled for embedding generation by the system.


\textbf{Semantic Indexing: }Once text splitting was done, each chunk of text was converted into a dense vector embedding to express semantic meaning. Embeddings were created using the all-miniLM-L6-v2 \footnote{https://huggingface.co/sentence-transformers/all-MiniLM-L6-v2} model that yields 384-dimensional vectors optimized for semantic similarity tasks. The vectors were then put into a Chroma vector store for storage due to its extreme lightness, ability to persist data, and fast similarity search mechanisms. Persisting the embeddings saved the system from recomputing the embeddings across sessions, reducing latency. The embeddings clustered semantically similar text chunks together in vector space, thereby allowing the retrieval of pertinent knowledge at query time.


\begin{figure*}
     \centering
     \begin{subfigure}[b]{0.76\textwidth}
         \centering
         \includegraphics[width=\textwidth]{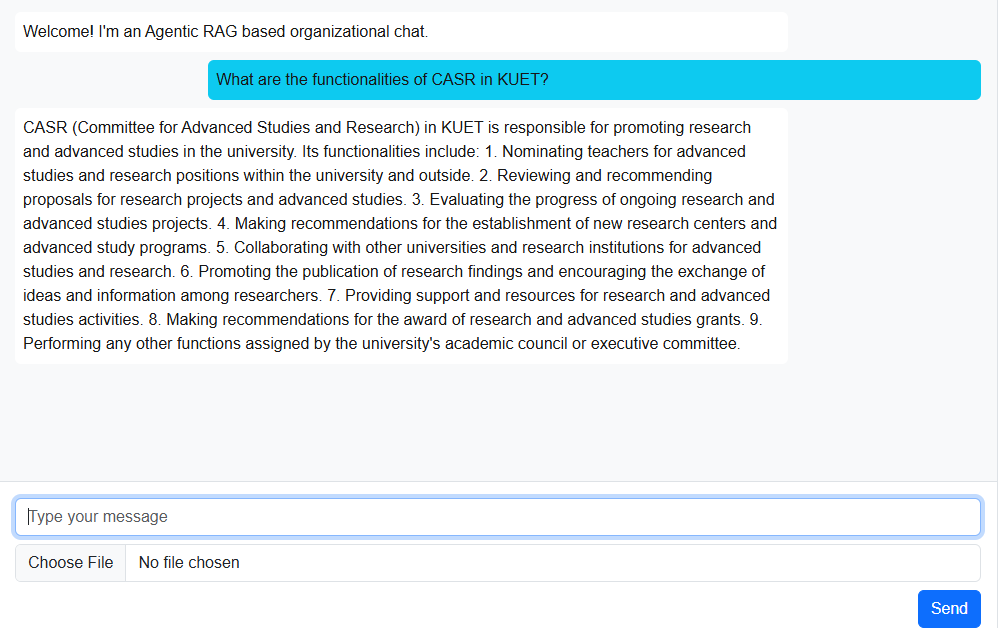}
         \caption{text only}
         \label{text_simulation}
     \end{subfigure}
     \begin{subfigure}[b]{0.76\textwidth}
         \centering
         \includegraphics[width=\textwidth]{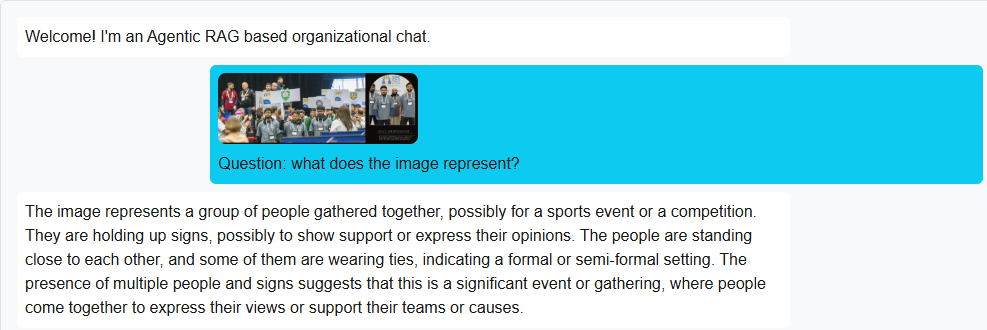}
         \caption{image with text}
         \label{image_simulation}
     \end{subfigure}
        \caption{Simulation of our system.}
\label{simulation}
\end{figure*}

\subsubsection{Retrieval}
The retriever acts as a bridge between user queries and the knowledge base itself. When the query is submitted, it must be embedded into the same semantic space using the same embedding model to maintain uniformity in representation. Then, the query embedding is contrasted with the stored document embeddings by their cosine similarity to select the highly relevant chunks. Here, the top-3 were retrieved, as a good balance between relevance and efficiency was considered. 

\subsubsection{Augmentation via Context-Query Fusion}
After retrieving the most relevant top-3 chunks based on the user’s query, concatenates these segments to create one contextual block. This contextual block is then joined with the original query to create an augmented prompt, which is then passed to the language model. The final prompt template is as follows: 
\begin{quote}
``Use the following university-verified information to answer. If the information is insufficient, say `I cannot answer the question.'\\
Retrieved Context 1: \dots\\
Retrieved Context 2: \dots\\
Retrieved Context 3: \dots\\
User Query: \dots''
\end{quote}


Such prompt construction forces the model to be aware of domain-specific knowledge retrieved from the underlying corpus, thus sharpening its generation of grounded, context-aware responses, and reducing hallucinations.

\subsubsection{Performance Monitoring and Feedback Collection}
Feedback can be provided via a form that accepts details such as name, email, and comments on the outputs produced. This feedback is logged into a CSV file and later used for assessment and enhancement of the model.

\begin{figure*}
\centering
\centerline{\includegraphics[scale=0.6]{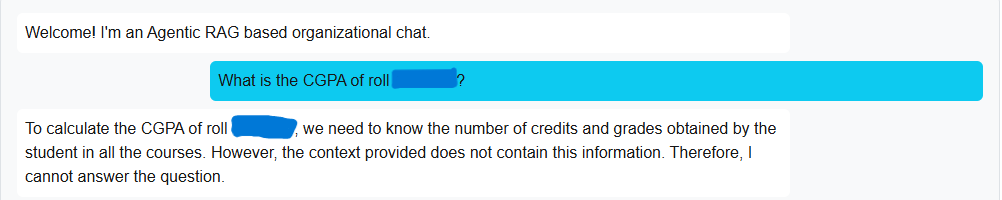}}
\caption{Demonstration of how our system handle hallucination.}
\label{handling_hallucination}
\end{figure*}

\begin{figure}
\centering
\centerline{\includegraphics[scale=0.4]{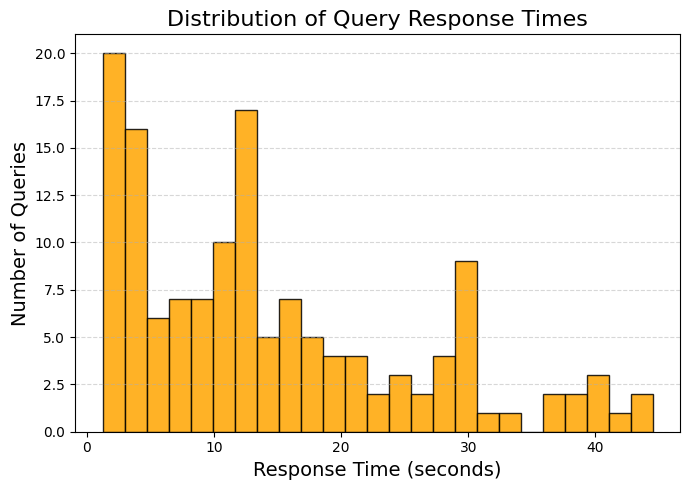}}
\caption{Distribution of query response time during the trial phase.}
\label{query_response}
\end{figure}



\subsection{Deployment}
The proposed chat assistant is publicly deployed at \href{https://chat.kuet.ac.bd}{\texttt{chat.kuet.ac.bd}} on an Intel Core i7-7700 CPU at 3.0 GHz (four CPUs), with 32 GB of main memory, and an NVIDIA GeForce GTX~1080 (8GB), running with Python 3.12.




\section{Experimental setup and simulation}
\label{sec:result}
In this section, we discuss the technologies that we have used together with the simulation of our system.

\subsection{Experimental setup}
The proposed system is built using a Next.js front end and a FastAPI back end. For visual–language processing, it employs the LLaVA-1.5-7B model, while ChromaDB serves as the vector database. The generative AI components are implemented using PyTorch. Documents are ingested through the PyPDFLoader module and segmented using the RecursiveCharacterTextSplitter. Semantic similarity search is used as the retrieval mechanism, supported by the all-MiniLM-L6-v2 embedding model. The overall retrieval-augmented generation (RAG) workflow is orchestrated through the LangChain framework.


\subsection{Interactive User Interface}
In Fig. \ref{base_ui}, the interactive chat user interface of our system has been shown, where users can ask queries (text and/or image) and get their required information. Users can also provide their feedback for further improvement of our system using the feedback form shown in Fig. \ref{feedback_form}.


\subsection{Simulation}
The simulation of our system is demonstrated in Fig. \ref{simulation}. In Fig. \ref{text_simulation}, the user asks a domain-specific query (KUET-related) only in text, and our RAG-based system is capable of giving a context-aware reply. And the user can also make a query based on an uploaded image, and our system is also capable of answering such a query, which is shown in Fig. \ref{image_simulation}.


\subsection{Handling Hallucination}
Fig. \ref{handling_hallucination} shows how our system can handle the LLM hallucination problem. Hallucination means that LLMs give wrong answers confidently, although there is no evidence \cite{liu2024survey}. It is a very common limitation of a general LLM models.However, as we are incorporating context and no context related to the query was found, the LLM answers that it doesn't have the information instead of giving a wrong answer.


\subsection{Query Response Time}
Fig. \ref{query_response} illustrates the distribution of query response times recorded during the trial, based on 160 queries. The response times range from 1.23 to 44.49 seconds, with a mean of 13.57 seconds and a standard deviation of 10.87 seconds. The histogram reveals a right-skewed distribution, indicating that most queries were answered quickly, but some took significantly longer. Despite GPU acceleration, response times remain significantly high because the model with 7 billion parameters requires substantial computation. The GTX 1080 can handle it, but it is not the best choice for large-scale inference. This limitation in deployment plays a crucial role in the latency experienced, and hence, the response time is longer. Also, the CPU-based preprocessing and the vector searches take up more time. Moreover, sequential processing lowers the usage of the GPU.

\subsection{Multimodal Performance Evaluation}
The feedback form filled out by 52 trial participants yielded a satisfaction score of 4.26/5 on average, indicating improved access to academic resources. Besides that, we also analyzed user satisfaction with the different query modalities. The system managed to reach the average score of 4.56/5 for text-only queries and 3.95/5 for image-based queries. The average response time for image queries was 16.4 seconds, about 4.6 seconds longer than for text-only ones, which was caused by the overhead of multimodal processing.




\subsection{RAG Effectiveness Analysis}
\begin{figure}
\centering
\centerline{\includegraphics[scale=0.5]{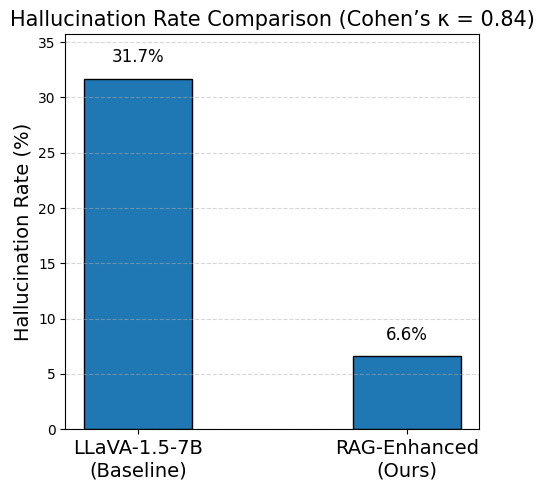}}
\caption{Demonstrating how our RAG-enhanced approach reduces baseline hallucination rate by about 25\%.}
\label{hallucination_rate}
\end{figure}

To provide quantitative evidence of system effectiveness, we evaluated the hallucination rate, retrieval accuracy, user satisfaction, and latency. Based on 160 domain-specific queries during the trial phase, the baseline LLaVA-1.5-7B model exhibited a hallucination rate of 31.7\%, whereas our RAG-enhanced model reduced this to 6.6\%, which is visually illustrated in Fig. \ref{hallucination_rate}. Hallucination annotations were validated independently by two reviewers, achieving a substantial inter-annotator agreement with Cohen’s $\kappa$ = 0.84, confirming the reliability of the evaluation \cite{warrens2015five}. The semantic retriever achieved 78.3\% top-1 and 92.1\% top-3 accuracy. As the top-3 were used in our system, retrieval accuracy remains high for most university-related queries.




\section{Data Privacy, Secure Access Control, and Compliance}
\label{sec:privacy}
Institutional documents and user queries are dealt with under very strict security measures. Retrieval functions involve only JWT-based authorization, and the entire communication of the client and server is protected with HTTPS. Sensitive data like emails or IDs are already masked before logging, and uploaded images are processed only in memory unless permission is granted. Furthermore, very strict CORS policies are in place that allow requests from only the trusted university domains. Not only do these actions assure privacy, but they also comply with the institutional data governance policies.



\section{Conclusions}
\label{sec:conclusion}
The study shows the effectiveness of combining RAG with multimodal inference to create a smart university chat assistant that can handle both text and image queries. By using semantic embeddings, vector storage, and retrieval pipelines, the system provides responses based on context. This reduces the reliance on static knowledge bases and helps avoid the hallucinations often seen in large language models. The chat assistant is deployed as a publicly available web application, demonstrating its practical use for various university stakeholders. This method emphasizes how domain-specific, retrieval-augmented chat assistants can make information more accessible, streamline administrative tasks, and help close digital gaps in educational institutions, especially in developing areas. Though our system answers most of the domain-specific questions correctly, there are still several limitations, like multihop reasoning, latency issues, etc. Sometimes we also face the contextual integration problem, thus resulting in a very generic response. In the future, we would like to extend our system to be more reliable using advanced RAG variants like agentic RAG, multi-hop RAG, etc.  

\bibliographystyle{./IEEEtran}
\bibliography{./IEEEexample_rag}

\end{document}